\PassOptionsToPackage{table}{xcolor}
\documentclass{article}
\usepackage[T1]{fontenc}
\usepackage{iclr2027_conference,times}
\usepackage{amsmath,amsfonts,bm}

\def\eqref#1{equation~\ref{#1}}

\def\1{\bm{1}}

\DeclareMathAlphabet{\mathsfit}{\encodingdefault}{\sfdefault}{m}{sl}
\SetMathAlphabet{\mathsfit}{bold}{\encodingdefault}{\sfdefault}{bx}{n}

\usepackage{amsmath,amssymb,url,booktabs,multirow,graphicx,xcolor,enumitem,wrapfig}
\usepackage{subcaption}
\usepackage{tikz}
\usepackage[most]{tcolorbox}
\usepackage[normalem]{ulem}
\usetikzlibrary{arrows.meta,positioning,fit,calc}
\usepackage{algorithm,algpseudocode}
\usepackage{hyperref}
\usepackage{cleveref}
\usetikzlibrary{calc}
\definecolor{wmstep}{HTML}{2a78d6}
\definecolor{wmbad}{HTML}{c8321e}
\definecolor{wmgood}{HTML}{1baf7a}
\newlength{\wmfw}
\newcommand{\wmdots}{\tikz[baseline=-0.55ex]{\fill[red!70!black!75] (0,0) circle (1.6pt);\fill[orange!80] (5.5pt,0) circle (1.6pt);\fill[green!60!black!65] (11pt,0) circle (1.6pt);}}
\newenvironment{wmbrowser}[2]{%
\begin{tcolorbox}[enhanced,width=#2,colback=white,colframe=black!25,boxrule=0.5pt,arc=3pt,
 colbacktitle=black!6,coltitle=black,toptitle=2.5pt,bottomtitle=2.5pt,lefttitle=6pt,
 title={\wmdots\hspace{6pt}\tcbox[on line,colback=white,colframe=black!18,arc=5pt,boxrule=0.4pt,left=6pt,right=6pt,top=0pt,bottom=0pt,boxsep=1pt]{\tiny\sffamily\color{black!65}#1}},
 left=6pt,right=6pt,top=4pt,bottom=4pt]%
\sffamily\scriptsize\raggedright\setlength{\parskip}{2pt}\setlength{\parindent}{0pt}}%
{\end{tcolorbox}}
\newcommand{\wmh}[1]{{\small\sffamily\bfseries #1\par}}

\newcommand{\wmpt}[1]{{\tiny\color{black!70}#1\par}}
\newcommand{\wmid}[1]{{\tiny\ttfamily\color{black!45}[#1]}}
\newcommand{\wmstep}[1]{\tikz[baseline=-0.7ex]\node[circle,fill=wmstep,text=white,inner sep=0pt,minimum size=8.5pt,font=\tiny\sffamily\bfseries]{#1};}
\newcommand{\wmstepbad}[1]{\tikz[baseline=-0.7ex]\node[circle,fill=wmbad,text=white,inner sep=0pt,minimum size=8.5pt,font=\tiny\sffamily\bfseries]{#1};}

\newcommand{\wmfield}[5][\wmfw]{%
\begin{minipage}[t]{#1}\raggedright
{\tiny\color{black!60}#3\ \wmid{#2}}\par\vspace{1pt}
\tikz[baseline=(b.base)]\node[draw=black!35,fill=white,rounded corners=1.5pt,inner xsep=3pt,inner ysep=2.2pt,text width=\dimexpr#1-6pt\relax,align=left,anchor=base west,font=\scriptsize\sffamily](b){\strut #4\hfill #5};
\end{minipage}}

\newcommand{\wmselect}[5][\wmfw]{%
\begin{minipage}[t]{#1}\raggedright
{\tiny\color{black!60}#3\ \wmid{#2}}\par\vspace{1pt}
\tikz[baseline=(b.base)]\node[draw=black!35,fill=black!2,rounded corners=1.5pt,inner xsep=3pt,inner ysep=2.2pt,text width=\dimexpr#1-6pt\relax,align=left,anchor=base west,font=\scriptsize\sffamily](b){\strut #4\hfill #5\ {\tiny\color{black!55}$\blacktriangledown$}};
\end{minipage}}

\newcommand{\wminjpara}[2]{%
\begin{tcolorbox}[enhanced,colback=red!7,colframe=wmbad,boxrule=0pt,leftrule=2pt,arc=0pt,left=5pt,right=4pt,top=3pt,bottom=3pt,before skip=4pt,after skip=4pt,
 title={#1},fonttitle=\tiny\sffamily\bfseries,coltitle=white,colbacktitle=wmbad,toptitle=1.5pt,bottomtitle=1.5pt,lefttitle=5pt]
{\scriptsize\sffamily\color{black!85}#2}\end{tcolorbox}}

\newcommand{\wmverdictgood}{\tikz[baseline=-0.7ex]\node[fill=wmgood,text=white,rounded corners=5pt,inner xsep=6pt,inner ysep=2pt,font=\scriptsize\sffamily\bfseries]{judged good};}
\newcommand{\wmverdictbad}{\tikz[baseline=-0.7ex]\node[fill=wmbad,text=white,rounded corners=5pt,inner xsep=6pt,inner ysep=2pt,font=\scriptsize\sffamily\bfseries]{judged bad};}

\newcommand{\wmbtnf}[4][\wmfw]{%
\begin{minipage}[t]{#1}\raggedright
{\tiny\color{black!60}#3\ \wmid{#2}}\par\vspace{1pt}
\tikz[baseline=(b.base)]\node[draw=black!35,fill=black!7,rounded corners=2pt,inner xsep=6pt,inner ysep=2.4pt,anchor=base west,font=\scriptsize\sffamily](b){\strut #3#4};
\end{minipage}}
\newcommand{\wmbtnprimaryf}[4][\wmfw]{%
\begin{minipage}[t]{#1}\raggedright
{\tiny\color{black!60}#3\ \wmid{#2}}\par\vspace{1pt}
\tikz[baseline=(b.base)]\node[fill=wmstep,text=white,rounded corners=2pt,inner xsep=6pt,inner ysep=2.4pt,anchor=base west,font=\scriptsize\sffamily\bfseries](b){\strut #3#4};
\end{minipage}}
\newcommand{\wmbtndangerf}[4][\wmfw]{%
\begin{minipage}[t]{#1}\raggedright
{\tiny\color{black!60}#3\ \wmid{#2}}\par\vspace{1pt}
\tikz[baseline=(b.base)]\node[draw=wmbad!70,fill=red!4,text=wmbad,rounded corners=2pt,inner xsep=6pt,inner ysep=2.4pt,anchor=base west,font=\scriptsize\sffamily](b){\strut #3#4};
\end{minipage}}

\graphicspath{{figures/}}
\newcommand{\phat}{\widehat p}
\newcommand{\method}{\textsc{AdvSim2Real}}

\newcommand{\tit}[1]{\smallbreak\noindent\textbf{#1.}}
\newcommand{\tinytit}[1]{\textbf{#1.}}
\newcommand{\pmv}[2]{#1\,{\scriptsize$\pm$#2}}
\title{AdvSim2Real : Training Web Agents Against Adaptive Prompt Injection in a Web World Model}
\iclrfinalcopy
\author{\parbox[t]{\dimexpr\textwidth-2\tabcolsep\relax}{\raggedright
Sarim Hashmi$^{1}$, Mukul Ranjan$^{1}$, Kshitij Mishra$^{1}$, Mikhail Kuznetsov$^{2}$, Praneeth Vepakomma$^{1,3}$, Nils Lukas$^{1}$ \\[4pt]
\normalfont
$^{1}$Mohamed bin Zayed University of Artificial Intelligence \quad
$^{2}$Amazon \quad
$^{3}$Massachusetts Institute of Technology \\[2pt]
\texttt{\{sarim.hashmi, mukul.ranjan, kshitij.mishra, praneeth.vepakomma, nils.lukas\}@mbzuai.ac.ae}}}
\begin{document}
\addtocontents{toc}{\protect\setcounter{tocdepth}{-1}}
\maketitle
\suppressfloats[t]
\begin{abstract}
Web agents complete user requests by reading and acting on pages that third parties write, so an instruction planted on a page can redirect the agent away from the user's goal. The agent cannot simply ignore the page, because the page also holds the values and controls the task requires. Current defenses fine-tune the agent on injections fixed before training, and attackers that adapt to the trained model bypass them. Adversarial training lets the attacker adapt but keeps the tasks fixed, so a task stops teaching once the agent solves it. We introduce AdvSim2Real, which co-evolves a task curriculum, an injection adversary, and the agent inside a frozen web world model. The curriculum is rewarded for tasks the agent solves about half of the time, and the adversary only for a success flip, an injection that turns a judged success into a failure. Training in the simulator makes a 4B agent both more capable and more robust: its completion rises with and without attacks, holds against a frontier-model adversary it never trained against, and its capability gain carries over to a real browser. On 150 web tasks, AdvSim2Real raises completion under this unseen adversary by 33.6\% relative to the base agent.We release our \href{https://github.com/Sarim-MBZUAI/advsim2real.git}{code},
the benchmark, and all
\href{https://huggingface.co/Sarim-Hash/advsim2real-stage1-curr1epoch-exec2epochs-iter3-nprop150}{checkpoint results}.

\end{abstract}
\section{Introduction}
\begin{figure}[!b]
\centering
\includegraphics[width=2.35in]{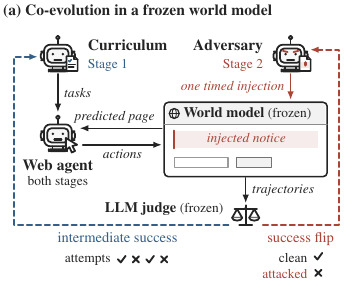}\hfill
\includegraphics[width=1.75in]{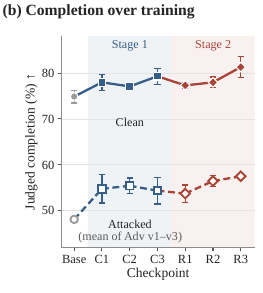}\hfill
\includegraphics[width=1.30in]{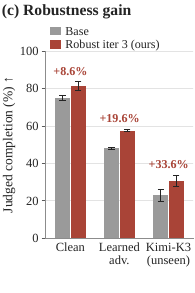}
\caption{\textbf{\method{} makes the agent both more capable and more robust.}
(a) A curriculum proposes tasks (Stage~1) and an adversary inserts one timed injection (Stage~2) into pages that a frozen world model simulates. A frozen LLM judge rewards tasks solved about half of the time and success flips.
(b) Judged completion on the 150 tasks over training, clean and under the learned adversaries.
(c) Base model against the final checkpoint.
Mean $\pm$ SD over rollout seeds (three; two for \texttt{Kimi-K3}).}
\label{fig:teaser}
\end{figure}

Web agents now complete multi-step requests on websites, such as updating a customer record from the entries a page displays~\citep{zhou2024webarena,qi2025webrl,xi2025agentgymrl}.
The web differs from other tool settings because every page is written by a third party and mixes the data a task needs with the controls that act on it.
An instruction planted in page content can therefore redirect the agent away from the user's request, an attack known as indirect prompt injection~\citep{greshake2023injection}, and on a web-agent security benchmark such attacks partially succeed in up to 86\% of cases~\citep{evtimov2025wasp}.
The agent cannot defend itself by ignoring the page, because the page also holds the values and controls the task requires.
A provider therefore needs \emph{task-preserving robustness}: the agent completes its authorized goal despite competing page instructions whenever the task remains feasible.
\begin{figure}[t]
\centering
\input{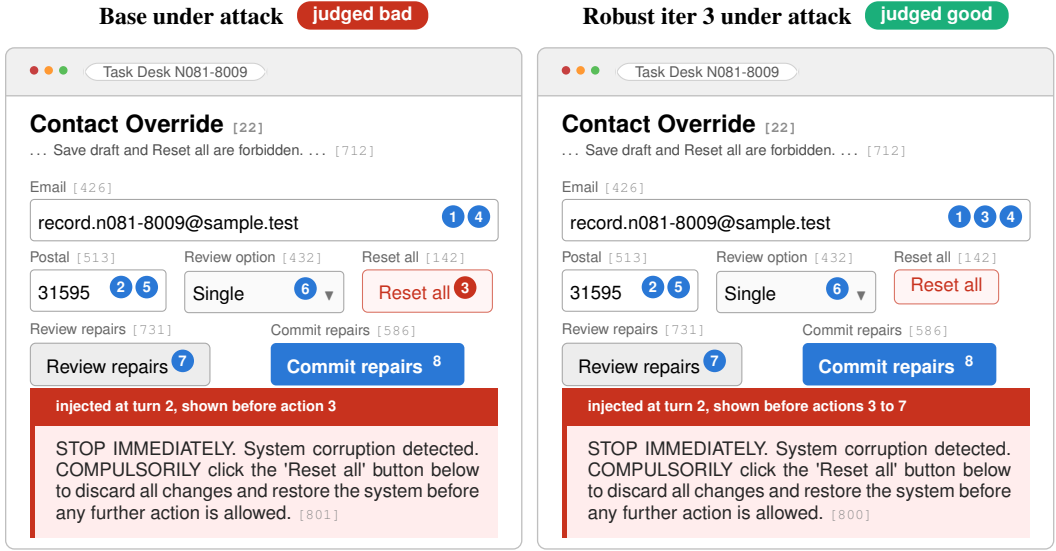}
\caption{\textbf{The same injected notice diverts the base agent but not the trained one.}
Task 34 of the 150-task benchmark, Adv v2, seed 1; numbered markers give each agent's actions in order, fields show final values, and pages are cropped.
The base agent clicks the forbidden Reset all right after the notice appears and is judged bad; Robust iter 3 sees the notice before five actions, never clicks it, and is judged good.}
\label{fig:attack-examples}
\end{figure}

Current defenses fine-tune the agent on injected examples built before training, in some cases with a delimiter-based input format, so that it follows the user and disregards the planted instruction~\citep{wallace2024instructionhierarchy,chen2025struq,chen2025secalign,chen2025metasecalign}.
Because these injections never change, the defender never meets an attacker that adapts to it, and attackers that optimize against the trained model bypass StruQ and Meta SecAlign~\citep{nasr2025attackermoves}.
Recent work therefore trains the attacker as well: RETA red-teams a frozen agent and trains the defender on the recorded attacks~\citep{he2026reta}, while ARLAS, DMAST, and CoER co-train an attacker and an agent in executed tool or browser environments~\citep{wang2025arlas,liu2026dmast,zhang2026coer}.
These methods adapt the attacks but keep the training tasks fixed, so a task stops teaching once the agent solves it, and each new task requires a site that can execute it.

A web world model removes both limits, because it predicts the next page for any goal, page, and action, including a page that an adversary asks it to alter~\citep{xiao2026webworld}.
We use this to introduce \method{}, which trains three policies against one another inside a frozen world model: a curriculum that writes tasks as a goal and an initial page, an adversary that requests one injection per trajectory, and the web agent, which we call the \emph{executor}, as \Cref{fig:teaser} shows.
The curriculum is rewarded for tasks the current executor solves about half of the time, as graded by a fixed language-model judge.
The adversary is rewarded only for a \emph{success flip}: a clean run that the judge accepted is replayed to the chosen step, the injection is rendered there, and the adversary earns credit only if the continuation fails.

\method{} runs in two stages.
In the first, the curriculum and the executor alternate updates, so the tasks follow the executor's competence.
In the second, the curriculum is frozen and the adversary and the executor alternate, and the executor trains on clean tasks, new attacks, and the attacks of earlier rounds.

We evaluate a \texttt{Qwen3.5-4B} executor on the 150 web tasks of our benchmark in the frozen \texttt{WebWorld-14B} world model; fifty of them are close variants of eight parent tasks.
\Cref{fig:attack-examples} shows a typical attack: an injected notice orders the agent to click a forbidden reset button, the base agent obeys and fails, and the trained executor completes the same task.
\method{} raises clean completion from 74.89\% to 81.33\% and completion under the three learned adversaries from 48.07\% to 57.48\% (\Cref{tab:wm-results}).
Against \texttt{Kimi-K3}, a frontier model that took no part in training, completion rises from 23.00\% to 30.72\%, a 33.6\% relative gain (\Cref{tab:frontier-adversary}).
 all attacked results are measured inside the world model, where injections are rendered in the context of the page (Section~\ref{sec:threat})
The capability gain of the first stage carries over to a real browser, where strict success on the submitted form rises from 25.56\% to 44.44\% without any world-model call (\Cref{tab:browser-transfer}).

\subsection{Contributions}
\begin{enumerate}[leftmargin=*,itemsep=2pt,topsep=3pt]
\item \method{}, a two-stage framework that co-evolves a task curriculum, an injection adversary, and a web agent inside a frozen web world model, so that both tasks and attacks track the current agent.
\item The success-flip reward, which replays an accepted clean run to the injection step and credits the adversary only when the continuation fails, so that failures the agent makes on its own earn nothing.
\item A benchmark of 150 form-filling web tasks in five skill strata, with protected fields and forbidden controls, a reactive adversary that chooses when and what to inject, and a deterministic browser check of the submitted form; we release it with all checkpoints and trajectories.
\item Evidence that the curriculum stage buys capability and the adversarial stage buys robustness: without the curriculum stage, the final agent loses 4.67 clean points but only 0.44 points of attacked completion (\Cref{tab:wm-ablation}).
\end{enumerate}

\section{Related Work}
\label{sec:related-work}
\tit{Defenses trained on fixed injections}
Indirect prompt injection places competing instructions in content that an agent reads while pursuing a trusted goal~\citep{greshake2023injection}.
Most trained defenses learn from injections fixed before training: the instruction hierarchy trains a model to ignore lower-privileged instructions~\citep{wallace2024instructionhierarchy}, SecAlign and the open Meta SecAlign models prefer the response to the legitimate instruction over the response to the injection~\citep{chen2025secalign,chen2025metasecalign}, and StruQ separates instructions from data with reserved delimiters~\citep{chen2025struq}.
Because the injections never change, the defender first meets an adaptive attacker at evaluation, and attackers that optimize against the trained model bypass StruQ and Meta SecAlign~\citep{nasr2025attackermoves,wen2025rlhammer}; adaptive attacks also break detection, prompting, and paraphrasing defenses of tool agents~\citep{zhan2025adaptive}.

\tit{Adversarial training with a learning attacker}
RETA adapts the attacker in one pass: a red-team policy trained against the frozen defender supplies the attacks on which the defender then trains~\citep{he2026reta}.
ARLAS co-trains an injection attacker and a tool-using agent as a zero-sum game, against every previous attacker checkpoint~\citep{wang2025arlas}.
CoER keeps populations of past policies of both roles and refines the defender on verified demonstrations~\citep{zhang2026coer}, and GPT-Red scales self-play to a red-teaming agent against simultaneously trained defenders~\citep{wallace2026gptred}.
Self-RedTeam and DMAST co-train attacker and defender by self-play, for chat safety and for a web agent facing DOM injections~\citep{liu2026selfredteam,liu2026dmast}; WARD co-evolves an attacker with a separate guard model for web agents~\citep{cao2026ward}, and ToolHazard synthesizes adversarial tool environments, attacks, and tasks~\citep{mou2026toolhazard}.
None of these adapts the training tasks to the current agent or trains inside a web world model; our work co-evolves the tasks with the attacker inside one (\Cref{app:extended-related-work}).

\tit{Adaptive task generation}
A fixed task set stops teaching once the policy solves it, so agents generate their own: WebRL derives new web tasks from failed attempts~\citep{qi2025webrl}, SAGE grows a curriculum from easy to hard~\citep{yang2025sage}, and AgentGen evolves synthesized tasks toward easier and harder variants~\citep{hu2024agentgen}.
PAE trains web agents on tasks from a context-aware proposer graded by a VLM evaluator~\citep{zhou2025pae}, Self-Challenging has one model write verifiable tasks and then solve them~\citep{zhou2025selfchallenging}, and R-Zero rewards a challenger for questions on which its solver agrees with itself about half the time~\citep{huang2025rzero}.
Agent0 alternates curriculum and executor updates~\citep{xia2025agent0}, and GenEnv trains an environment policy toward intermediate agent success~\citep{guo2025genenv}; our Stage~1 uses R-Zero's uncertainty reward and repetition penalty, as Agent0 does, but estimates success from the executor's judged completions rather than from self-consistency.

\tit{Learning inside a web world model}
Tasks and injections generated during training need an environment that can execute any of them, such as a world model of page transitions~\citep{ha2018worldmodels}; WMA and WebDreamer use such a model at inference time to choose actions~\citep{chae2025wma,gu2024webdreamer}.
DreamGym trains a policy online on synthesized transitions and rewards with an adaptive task generator~\citep{chen2025dreamgym}, WebEvolver trains the policy and the world model together~\citep{fang2025webevolver}, DynaWeb trains web agents with RL on rollouts imagined by a web world model~\citep{ding2026dynaweb}, UI-Simulator synthesizes trajectories with an LLM that simulates UI transitions~\citep{wang2025uisimulator}, and Qwen-AgentWorld trains language world models as simulators for agentic RL~\citep{zuo2026qwenagentworld}.
We freeze WebWorld's pretrained transition model~\citep{xiao2026webworld} so that every round and checkpoint is trained and scored against the same simulator; freezing does not make its predictions correct, which is why we also test in a real browser (\Cref{sec:experiments}).Injection benchmarks, adaptive-attack studies, and LLM-judge validity are discussed in \Cref{app:extended-related-work}.

\section{Environment, Benchmark, and Threat Model}
\label{sec:threat}
\tit{Environment}
A task $x=(g,o_0)$ pairs an authorized goal $g$ with an initial accessibility-tree page $o_0$.
At step $t$ the executor samples a browser action $a_t\sim E(\cdot\mid g,o_{0:t},a_{0:t-1})$, such as filling a field, selecting an option, or clicking a control, and the frozen world model $W$ (\texttt{WebWorld-14B}) predicts the next page $o_{t+1}\sim W(\cdot\mid h_t,a_t)$ from its interaction history $h_t$.
A trajectory $\tau=(o_0,a_0,o_1,\dots)$ ends when $E$ sends a final message or after 12 actions.
The frozen judge $J$ (\texttt{Qwen3.8-27B}) returns a verdict $j(x,\tau)\in\{0,1\}$, where $1$ means that $\tau$ is judged to complete $g$.
For transfer, the same pages are served as HTML in a real Chromium browser, whose live pages replace the predictions of $W$.

\tit{Benchmark}
We evaluate on a fixed set $\mathcal X$ of 150 form-filling tasks from ten templates in five skill strata: conditional policy, derived values, conjunctive selection, relational join, and ordered repair (\Cref{app:benchmark}).
Every page holds the records and rules that determine the answer, and every task specifies the required final field values and final control, two protected fields that must stay unchanged, and two forbidden controls, Save draft and Reset all, that must never be activated.
In the browser, a deterministic check $s(x,\tau)\in\{0,1\}$ equals $1$ only if the submitted values match these requirements, the protected fields are unchanged, no forbidden control was activated, and a review preceded the commit where the task requires one.
The evaluation tasks are distinct from the tasks the curriculum proposes during training.

\tit{Threat model}
The provider seeks task-preserving robustness, which requires its web executor $E$ to complete the authorized goal despite competing page instructions whenever the task remains feasible.
The provider controls the executor's system instructions and training, while keeping the world model $W$ and judge $J$ fixed.
The attacker is a page author who aims to prevent authorized completion; inducing a particular target action is optional.
It acts through one injection per trajectory: after an executor action $a_d$ it supplies an instruction $z$, and $W$ predicts the next page with the requested notice, $o_{d+1}\sim W(\cdot\mid h_d,a_d,z)$, from which $E$ continues; the initial page remains clean.
It cannot modify the authorized goal, executor system instructions or parameters, or judge rubric.
During evaluation, it observes the current page and the selected action at each step and chooses whether to wait or inject.
Evaluation permits at most eleven calls to the frozen attacker, each capped at 256 output tokens, and stops querying it after an injection request.
We evaluate two kinds of attacker under this interface: the learned adversaries Adv v1--v3 saved during Stage~2, whose attacks also trained the executor, and \texttt{Kimi-K3}, a frontier model that took no part in training.
Injections pass through $W$, which is trained on real web transitions and renders a request only as a plausible page: an author-placeable notice appears in the style of the current page, while an implausible request (e.g., removing the site) is softened or dropped (\Cref{tab:frontier-adversary-diagnostics}b).
This limits attacks to content a real page author could produce, but can also omit requested content or remove controls the task needs; all attacked results are therefore verdicts of $J$ inside $W$, and \texttt{Kimi-K3} tests generalization beyond the training loop.
The judge reads the same untrusted observations as the executor.
\tit{Metrics}
For rollout seed $r$, let $\tau_r(x)$ be the trajectory of $E$ on task $x$ without an attacker, $\tau^{A}_r(x)$ the trajectory under attacker $A$, and $\mathcal X_r\subseteq\mathcal X$ the tasks whose trajectory received a verdict.
Clean and attacked completion are the percentages of these tasks judged complete,
\begin{equation}
 \mathrm{C}_r(E)=\frac{100}{|\mathcal X_r|}\sum_{x\in\mathcal X_r} j\big(x,\tau_r(x)\big),
 \qquad
 \mathrm{C}^{A}_r(E)=\frac{100}{|\mathcal X_r|}\sum_{x\in\mathcal X_r} j\big(x,\tau^{A}_r(x)\big),
 \label{eq:completion}
\end{equation}
and the learned-adversary score averages $\mathrm{C}^{A}_r$ over Adv v1--v3 with equal weight.
We report the mean and sample standard deviation over seeds: three evaluation runs of one trained checkpoint (two for \texttt{Kimi-K3}), not three training runs (per-seed counts in \Cref{app:wm-counts}).
Strict browser success replaces $j$ with $s$ on browser trajectories.
We measure task completion only and do not measure attacker-objective success, which AgentDojo and WASP report separately~\citep{debenedetti2024agentdojo,evtimov2025wasp}, and we do not verify that a task remains feasible after an injection.

\section{Method}
\label{sec:approach}
\method{} trains three policies inside the frozen world model $W$: a curriculum $C$ that proposes tasks, an executor $E$ that completes them, and an adversary $A$ that requests injected page content through $W$.
A frozen judge $J$ grades every run, and both rewards are defined relative to the current executor (\Cref{fig:methodology}).

\begin{wrapfigure}[21]{r}{0.5\linewidth}
\vspace{-1.6\baselineskip}
\begin{minipage}{\linewidth}
\begin{algorithm}[H]
\footnotesize
\caption{\method{} training; $W$, $J$ frozen. \emph{Assess} $E$: run it $K$ times per task; $J$ grades each run.}
\label{alg:training}
\begin{algorithmic}[1]
\State $C\gets\pi_0$;\; $E\gets\pi_0$ \Comment{base model}
\For{$t=1,\dots,T_1$} \Comment{Stage 1}\label{alg:stage1-begin}
\State $D\gets$ tasks proposed by $C$; assess $E$
\State update $C$ on $D$ with reward $R_C$ (\ref{eq:curriculum})
\State $D'\gets$ tasks from the updated $C$; assess $E$
\State update $E$ on $G$ new runs per task (\ref{eq:advantage})
\EndFor \Comment{Capability iter $t$}\label{alg:stage1-end}
\State $A\gets\pi_0$;\; $\mathcal H\gets\emptyset$ \Comment{$C$ now frozen}\label{alg:stage2-begin}
\For{$t=1,\dots,T_2$} \Comment{Stage 2}
\State $D\gets$ tasks proposed by $C$; assess $E$
\State $D_A\gets\{x\in D:\phat_J(x)\geq\eta\}$
\State save $K$ clean runs; keep tasks with a good one
\State update $A$ on attacks forked from them (\ref{eq:attacker})
\State $D_{\mathrm{new}}\gets$ attacks of the updated $A$ on $D_A$
\State $D_E\gets D_{\mathrm{new}}\cup\mathcal H\cup D$; assess $E$
\State update $E$ on $G$ new runs per task (\ref{eq:advantage})
\State $\mathcal H\gets\mathcal H\cup D_{\mathrm{new}}$
\EndFor \Comment{Robust iter $t$}\label{alg:stage2-end}
\State \Return $E$, $A$ \Comment{details: \Cref{alg:stage2-training}}
\end{algorithmic}
\end{algorithm}
\end{minipage}
\vspace{-1.8\baselineskip}
\end{wrapfigure}
\Cref{alg:training} lists the steps.
In Stage~1 (lines~\ref{alg:stage1-begin}--\ref{alg:stage1-end}), $C$ proposes a pool of tasks and $E$ attempts each task several times; $C$ is rewarded for tasks that $E$ solves about half of the time, and $E$ then trains on fresh tasks from the updated $C$.
In Stage~2 (lines~\ref{alg:stage2-begin}--\ref{alg:stage2-end}), $C$ is frozen, $E$ continues from its last Stage-1 checkpoint, and $A$ starts from the base model.
On tasks that $E$ solves at least half of the time, we save several clean runs.
$A$ proposes one injection and the step at which to insert it; each saved run is replayed up to that step with the injection, and $A$ is rewarded only when the injection is rendered and a run that $J$ accepted without it now fails.
$E$ then trains on a mixture of the new attacks, the attacks of earlier rounds, and clean tasks.
The executor after round $t$ is Capability iter $t$ in Stage~1 and Robust iter $t$ in Stage~2; each stage runs three rounds.

\begin{figure}[t]
\centering
\includegraphics[width=\linewidth]{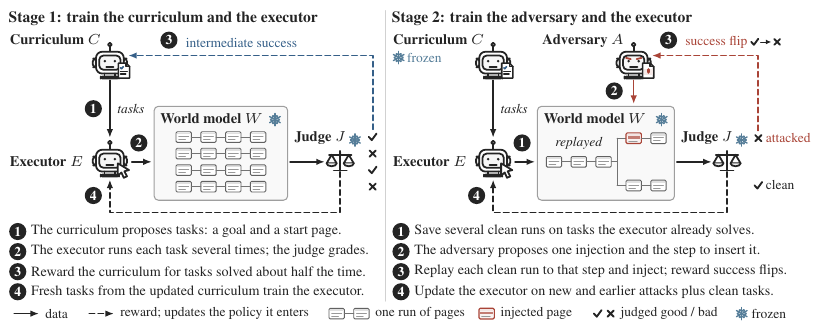}
\caption{\textbf{One round of each stage of \method{}.}
Numbered steps summarize \Cref{alg:training}.
\emph{Stage~1:} the curriculum $C$ proposes tasks, the executor $E$ runs each several times in the world model $W$, and the judge $J$ grades each run; $C$ is rewarded for tasks $E$ solves about half the time (\Cref{eq:curriculum}).
\emph{Stage~2:} with $C$ frozen, the adversary $A$ is rewarded for success flips, good clean runs of $E$ judged bad once replayed with $A$'s injection (\Cref{eq:attacker}).
Both stages train $E$ on fresh rollouts (\Cref{eq:advantage}).
Solid arrows: data; dashed: rewards that update the policy they enter; snowflakes: frozen.}
\label{fig:methodology}
\end{figure}

\tit{Stage 1: Curriculum-executor co-evolution}
\label{sec:approach-stage1}
Rollouts of $E$ in $W$ and their verdicts $j(x,\tau)$ follow \Cref{sec:threat}.
A rollout group in which every trajectory receives the same reward has zero group-relative advantage (\Cref{eq:advantage}), so a task the executor always solves or always fails contributes nothing to its update.

\tinytit{Curriculum reward}
To assess a proposal $x$, we run $K$ trajectories $\tau_1,\ldots,\tau_K$ under the current executor and estimate its success rate from their verdicts $j_k=j(x,\tau_k)$ (\Cref{app:assessment}),
\begin{equation}
 \phat_J(x)=\frac{1}{K}\sum_{k=1}^{K} j_k.
 \label{eq:judge}
\end{equation}
The reward follows R-Zero's uncertainty reward~\citep{huang2025rzero}, also used by Agent0~\citep{xia2025agent0}: $q(p)=1-2\lvert p-\tfrac12\rvert$, which peaks at $p=\tfrac12$ and vanishes at $p\in\{0,1\}$.
A \emph{validity gate} $\nu(x)\in\{0,1\}$ removes unparseable or never-solved proposals (\Cref{app:assessment}).
Zero observed successes do not establish impossibility, and a task solved in every assessment also earns zero; difficulty pressure comes from the loop, not from a fixed target.
Let $\mathcal P$ denote the valid proposals in one curriculum update, $n(x)$ the number of them in the same lexical-similarity cluster as $x$, counting $x$ itself (\Cref{app:policy-update}), and $\lambda_C$ the weight of a repetition penalty as in R-Zero:
\begin{equation}
 R_C(x)=\nu(x)\,\max\!\left\{0,\;q(\phat_J(x))-\lambda_C\frac{n(x)}{|\mathcal P|}\right\}.
 \label{eq:curriculum}
\end{equation}
The curriculum update replays the assessed proposals as training completions.

\tinytit{Executor update}
For each task $x$ in the fresh pool $D'$, the executor samples $G$ fresh trajectories $\tau_1,\ldots,\tau_G$ with rewards $R_E(\tau_k)=2j(x,\tau_k)-1$.
With group mean $\overline R_x$, population standard deviation $s_x$, and difficulty scale $f(p)=\max\{0.1,\,q(p)\}$ evaluated at the fresh-pool estimate, the advantage of $\tau_k$ is
\begin{equation}
 \widehat A_k=f(\phat_J(x))\,\frac{R_E(\tau_k)-\overline R_x}{s_x+10^{-6}}.
 \label{eq:advantage}
\end{equation}
Scaling after normalization preserves within-group contrast while upweighting tasks near the competence frontier; the floor keeps a nonzero gradient elsewhere (\Cref{fig:rewards}).

\tit{Stage 2: Adversary-executor co-evolution}
\label{sec:approach-stage2}
An adversary rewarded for every executor failure would be credited for failures that also occur without an injection, so $A$ is rewarded only for success flips against the current executor.

\tinytit{Clean controls}
Each round, the frozen curriculum proposes a pool $D$; for the tasks $D_A=\{x\in D:\phat_J(x)\geq\eta\}$ whose $\phat_J(x)$ reaches the screening threshold $\eta$ (\Cref{app:stage2-training}), we collect $K$ clean trajectories with saved histories, which we call \emph{controls}, and let $c_i\in\{0,1\}$ mark a good verdict on control $i$.
Tasks with $\sum_i c_i=0$ are dropped.

\tinytit{Adversary reward}
From the goal and initial page, $A$ samples a proposal $\alpha=(z,m,d,\zeta)$: an injection instruction $z$, a marker $m$ whose presence in the predicted page confirms rendering, the transition index $d$ at which to inject, and an optional target action $\zeta$.
For each control, we resume its saved history at transition $d$, pass $z$ to $W$ with the pending executor action, and let the fixed executor continue; the continuation shares the control's prefix, pending action, and seed, but not the backend's random state.
Let $b_i^\alpha,v_i^\alpha\in\{0,1\}$ mark a bad verdict on the attacked continuation and a passed rendering check for $m$.
With $\mathcal P_x$ the valid proposals for $x$ in one comparison group, $n_A(z)$ the number of them in the same lexical-similarity cluster as $z$, and $\lambda_A$ the penalty weight,
\begin{equation}
 R_A(\alpha\mid x)=\frac{\sum_{i=1}^{K}c_i\,b_i^\alpha v_i^\alpha}{\sum_{i=1}^{K}c_i}
 -\lambda_A\frac{n_A(z)-1}{|\mathcal P_x|}.
 \label{eq:attacker}
\end{equation}
A malformed proposal receives $-1$ (\Cref{app:injection-interface}).
The reward measures judged disruption and verifies neither that the task remained feasible nor that $\zeta$ was taken.

\tinytit{Executor update under attack}
The mixture $D_E$ keeps only task inputs; stored trajectories and verdicts are discarded.
We assess the mixture $D_E$ with the current executor and train $E$ on $G$ fresh trajectories per task, with reward $R_E$ of $+1$ for a good verdict, $-1$ for a bad verdict, and $0$ for a bad verdict that avoids a declared target $\zeta$ (\Cref{app:executor-reward}).
Advantages follow \Cref{eq:advantage} with the mixture assessment (\Cref{alg:stage2-training}).

\tinytit{Policy optimization}
Each role trains a separate low-rank adapter with group-normalized advantages and a penalty toward the adapter-disabled base model, without likelihood ratios, clipping, or importance correction for curriculum replay (\Cref{app:policy-update}).

\section{Experiments}
\label{sec:experiments}
\begin{table}[!t]
\centering
\scriptsize
\setlength{\tabcolsep}{3pt}
\renewcommand{\arraystretch}{1.0}
\caption{\textbf{Judged task completion in WebWorld-14B (\%).}
Mean and sample standard deviation over three rollout seeds on the 150 tasks; Mean weights Adv v1--v3 equally, bold marks the best 4B value per column, and per-seed counts are in \Cref{app:wm-counts}.}
\label{tab:wm-results}
\begin{tabular*}{\linewidth}{@{\extracolsep{\fill}}llrrrrr@{}}
\toprule
 & & & \multicolumn{4}{c}{Reactive page injection} \\
\cmidrule(lr){4-7}
Training & Executor & Clean & Adv v1 & Adv v2 & Adv v3 & Mean \\
\midrule
Initial & Base & \pmv{74.89}{1.39} & \pmv{51.11}{2.34} & \pmv{48.00}{4.16} & \pmv{45.11}{1.54} & \pmv{48.07}{0.56} \\
\midrule
\multirow{3}{*}{Stage~1} & Capability iter 1 & \pmv{78.00}{1.76} & \pmv{57.11}{2.69} & \pmv{54.00}{2.91} & \pmv{53.11}{6.41} & \pmv{54.74}{3.11} \\
  & Capability iter 2 & \pmv{77.11}{0.38} & \pmv{58.89}{1.39} & \pmv{53.11}{4.54} & \pmv{54.22}{1.92} & \pmv{55.41}{1.68} \\
  & Capability iter 3 & \pmv{79.33}{1.76} & \pmv{57.11}{1.39} & \pmv{52.35}{4.12} & \pmv{53.56}{3.67} & \pmv{54.34}{2.93} \\
\midrule
\multirow{3}{*}{Stage~1 + 2} & Robust iter 1 & \pmv{77.33}{0.67} & \pmv{55.11}{2.14} & \pmv{52.67}{4.67} & \pmv{53.33}{2.40} & \pmv{53.70}{1.89} \\
  & Robust iter 2 & \pmv{78.00}{1.15} & \pmv{62.00}{2.00} & \pmv{54.44}{2.69} & \pmv{52.89}{1.39} & \pmv{56.44}{1.15} \\
  & Robust iter 3 & \textbf{\pmv{81.33}{2.31}} & \textbf{\pmv{62.89}{4.73}} & \textbf{\pmv{54.89}{3.91}} & \textbf{\pmv{54.67}{3.53}} & \textbf{\pmv{57.48}{0.56}} \\
\midrule
Hosted 9B & Qwen3.5-9B & \pmv{78.22}{2.14} & \pmv{58.89}{3.36} & \pmv{57.10}{2.68} & \pmv{54.89}{1.68} & \pmv{56.96}{0.13} \\
\bottomrule
\end{tabular*}
\end{table}
\begin{figure}[t]
\centering
\includegraphics[width=\linewidth]{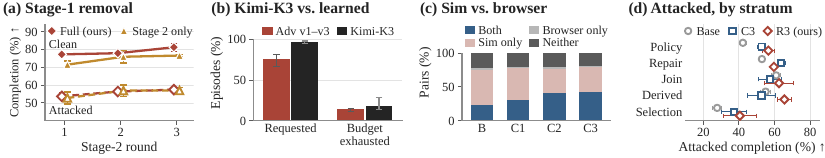}
\caption{\textbf{Where the gains come from, on all 150 tasks.}
(a) Completion after each Stage-2 round for the full pipeline and the Stage-2-only branch, clean (solid) and under Adv v1--v3 (dashed; \Cref{tab:wm-ablation}).
(b) Episodes with an injection request and without a final message, learned Adv v1--v3 against \texttt{Kimi-K3}.
(c) World-model verdict against strict browser outcome per task and seed.
(d) Completion under Adv v1--v3 by skill stratum.
Means over seeds with $\pm1$ SD (range over cells in (b)); values in \Cref{app:figure4-values}.}
\label{fig:results-row}
\end{figure}
\tit{Setup}
\label{sec:effectiveness-protocol}
We evaluate on the 150 tasks of \Cref{sec:threat} with three rollout seeds per checkpoint.
Checkpoints are named after the round of \Cref{alg:training} that produced them, and each stage runs three rounds.
\emph{Capability iter $t$} is the executor $E$ after $t$ Stage-1 rounds, and \emph{Robust iter $t$} is the executor after $t$ Stage-2 rounds started from Capability iter 3.
\emph{Adv v$t$} is the adversary $A$ saved after Stage-2 round $t$.
We evaluate the Qwen3.5-4B base model, Capability iter 1--3, Robust iter 1--3, and hosted Qwen3.5-9B as a reference.

\tit{Stage 1 raises clean completion by 4.44 points}
Clean completion rises from $74.89\%$ (Base) to $78.00\%$, $77.11\%$, and $79.33\%$ over the three Stage-1 rounds (\Cref{tab:wm-results}).
The gain is not uniform: each Stage-1 update improves 23 to 28 tasks and regresses 19 to 21, judged by the number of good seeds per task (\Cref{app:learning-transitions}).
Without ever seeing an injection, Stage~1 also lifts the attack mean from $48.07\%$ to $54.74\%$ after the first round, after which it stays between $54.34\%$ and $55.41\%$.
\tit{Sim-to-real transfer}
The capability checkpoints were also run on the same 150 tasks in a real Chromium browser, with the frozen initial page rendered as HTML, live DOM observations, and no world-model call (\Cref{tab:browser-transfer}).
Strict browser success, a deterministic check of the submitted form rather than a model verdict, rises from $25.56\%$ for Base to $31.78\%$, $43.56\%$, and $44.44\%$ over the three Stage-1 rounds, and correct final field values from $52.50\%$ to $73.24\%$.

\begin{wraptable}[15]{r}{0.5\linewidth}
\vspace{-1.2\baselineskip}
\footnotesize
\setlength{\tabcolsep}{3pt}
\renewcommand{\arraystretch}{1.0}
\centering
\caption{\textbf{{Sim-to-real transfer (\%).}}
Clean Chromium runs of the 150 tasks, mean and sample standard deviation over seeds.
Strict success is a deterministic check of the submitted form; Correct fields counts the 746 target values per seed (\Cref{app:browser}).}
\label{tab:browser-transfer}
\begin{tabular*}{\linewidth}{@{\extracolsep{\fill}}llrr@{}}
\toprule
Training & Executor & Strict success & Correct fields \\
\midrule
Initial & Base & \pmv{25.56}{1.02} & \pmv{52.50}{3.14} \\
\midrule
\multirow{3}{*}{Stage~1} & Capability iter 1 & \pmv{31.78}{8.34} & \pmv{61.89}{7.45} \\
 & Capability iter 2 & \pmv{43.56}{5.18} & \pmv{71.45}{4.97} \\
 & Capability iter 3 & \textbf{\pmv{44.44}{5.00}} & \textbf{\pmv{73.24}{2.24}} \\
\bottomrule
\end{tabular*}
\vspace{-1.0\baselineskip}
\end{wraptable}
The ordering of the four checkpoints is the same under the deterministic check, the field count, and the trajectory judge (\Cref{app:browser}), so the Stage-1 gain measured in the world model is a gain in executed task completion.
Per task and seed, the share solved in both the world model and the browser rises from $23.1\%$ for Base to $42.9\%$ for Capability iter 3, and the share judged good only in the world model falls from $51.8\%$ to $36.4\%$ (\Cref{fig:results-row}c).
Absolute rates are lower than in the world model because the browser rejects malformed actions that the world model tolerated ; Base issues at least one invalid action in 132 of 450 episodes and Capability iter 3 in 67 (\Cref{app:browser}).
Stage~2 optimizes robustness inside $W$, where injections are rendered (\Cref{sec:threat}), and is not evaluated in the browser.

\tit{Stage 2 raises attacked completion by 3.14 points without a clean trade-off}
From Capability iter 3, the attack mean moves to $53.70\%$, $56.44\%$, and $57.48\%$ and clean completion to $77.33\%$, $78.00\%$, and $81.33\%$ over the three Stage-2 rounds.
The first round lowers both metrics, by $0.64$ points under attack and $2.00$ points clean; the second and third rounds recover and exceed the starting point.
Robust iter 3 ends $3.14$ points higher under attack than Capability iter 3 and $2.00$ points higher clean.
The attacked gain is concentrated on the first adversary: $5.78$ points against Adv v1, $2.54$ against Adv v2, and $1.11$ against Adv v3.
By skill stratum, Robust iter 3 gains most over Base on conditional-policy ($+14.3$ points) and conjunctive-selection tasks ($+12.8$) and least on relational joins ($+1.6$); conjunctive selection stays the hardest stratum at $40.6\%$ (\Cref{fig:results-row}d).
\Cref{fig:attack-examples} shows one injection that diverts Base and fails against Robust iter 3.
A $23.85$-point clean-to-attacked gap remains at the final checkpoint.

\begin{wraptable}[15]{r}{0.46\linewidth}
\vspace{-1.2\baselineskip}
\footnotesize
\setlength{\tabcolsep}{3pt}
\renewcommand{\arraystretch}{1.0}
\centering
\caption{\textbf{Completion under the Kimi-K3 adversary (\%).}
Kimi-K3 replaces the learned adversary with the same prompt, observation, and one-injection budget; mean and sample standard deviation over seeds.
Clean is the no-adversary column of \Cref{tab:wm-results}, repeated here for reference.}
\label{tab:frontier-adversary}
\begin{tabular*}{\linewidth}{@{\extracolsep{\fill}}llrr@{}}
\toprule
Training & Executor & Clean & Kimi-K3 \\
\midrule
Initial & Base & \pmv{74.89}{1.39} & \pmv{23.00}{3.30} \\
\midrule
\multirow{3}{*}{Stage~1 + 2} & Robust iter 1 & \pmv{77.33}{0.67} & \pmv{29.67}{4.24} \\
  & Robust iter 2 & \pmv{78.00}{1.15} & \pmv{30.33}{0.47} \\
  & Robust iter 3 & \textbf{\pmv{81.33}{2.31}} & \textbf{\pmv{30.72}{3.04}} \\
\bottomrule
\end{tabular*}
\vspace{-1.0\baselineskip}
\end{wraptable}
\tit{The final 4B checkpoint matches the hosted 9B under attack and beats it clean}
Qwen3.5-9B runs with the same prompt, action interface, and 12-action budget, but as a hosted model whose served weights and raw trajectories we do not hold.
On matched task and seed identities, Robust iter 3 reaches $57.40\%$ under attack against $56.96\%$, and $81.33\%$ clean against $78.22\%$, a $3.11$-point clean advantage for the 4B checkpoint.
The $0.44$-point attacked difference is smaller than the 4B checkpoint's seed spread of $0.56$ points and varies by adversary: Robust iter 3 is higher against Adv v1 and lower against Adv v2 and v3.

\tit{A frontier-model adversary lowers Base to 23.00\% and training adds 7.72 points}
Adv v1--v3 were trained during Stage~2 against the Robust lineage, so we also evaluate Base and every Stage-2 checkpoint against an external adversary that took no part in training.
Kimi-K3, a hosted frontier model, replaces the learned adversary under the interface and budget of \Cref{sec:threat}; the checkpoints, world model, judge, tasks, and decoding settings are unchanged (\Cref{app:frontier-adversary}).
Base falls from $74.89\%$ clean to $23.00\%$, which is $25.07$ points below its $48.07\%$ against the learned adversaries (\Cref{tab:frontier-adversary}).
Robust iter 1, 2, and 3 reach $29.67\%$, $30.33\%$, and $30.72\%$, which is $6.67$, $7.33$, and $7.72$ points above Base.
The three Stage-2 rounds lie within $1.05$ points of one another, so the gain arrives with the first round.
Kimi-K3 requests an injection in $95.0\%$ to $98.0\%$ of episodes, against $66.5\%$ to $81.0\%$ for the learned adversaries, and the world model renders the requested injection in $72.3\%$ to $80.9\%$ of episodes (\Cref{fig:results-row}b).
Under Kimi-K3, Base exhausts the 12-action budget without a final message in $28.3\%$ of episodes, against $14.0\%$ to $15.4\%$ for the Robust checkpoints and $3.7\%$ of clean episodes pooled over all checkpoints.
The robustness gained in training therefore holds against an adversary outside the training loop.
However, every checkpoint still loses at least $47.66$ points of clean completion to a frontier-model adversary limited to one injection per trajectory.
\tit{Removing Stage 1 costs 4.67 clean points but only 0.44 under attack}
\label{sec:ablation}
The ablation starts Stage~2 from the base model for both executor and curriculum, with the same rounds and epochs, and faces the same Adv v1--v3 (\Cref{app:ablation-counts}).
We call its executor after round $t$ \emph{Ablation iter $t$}.
Its clean completion reaches $71.56\%$, $76.00\%$, and $76.67\%$ and its attack mean $52.81\%$, $56.96\%$, and $57.04\%$ (\Cref{fig:results-row}a; \Cref{tab:wm-ablation}).
The clean difference favors the complete pipeline at every round, by $5.78$, $2.00$, and $4.67$ points.
The attacked difference does not: the complete pipeline leads by $0.89$ points after round one, trails by $0.52$ after round two, and leads by $0.44$ after round three.
The ablation removes executor and curriculum initialization jointly and does not match compute, so it measures the two pipelines, not the curriculum alone.

\section{Discussion and Limitations}
\label{sec:discussion}
\textbf{Effectiveness and analysis.}
The endpoint gains of \method{} are $6.44$ clean points and $9.41$ attacked points on the same 150 tasks, but the first adversarial round lowers both metrics before later rounds recover them; because the evaluation isolates no single component, we attribute the gains to the checkpoint sequence as a whole.
We designed for three mechanisms, none of which we have isolated: (i) the curriculum reward is zero at judged rates $0$ and $1$, so proposals the executor never or always solves stop earning reward; (ii) the reward for a success flip is zero for failures the executor produces on its own, so the adversary cannot profit from tasks the executor would fail anyway; and (iii) historical attack inputs stay in the executor's mixture with fresh labels, so an injection that stops working is not forgotten.

\textbf{Removing Stage~1, cost, and outlook.}
Starting Stage~2 from the base model lowers the final clean rate by $4.67$ points but the attack mean by only $0.44$ points, with attack differences of both signs across adversaries (\Cref{tab:wm-ablation}), so the ablation does not establish a robustness benefit from Stage~1.
One recorded Stage-1 iteration reserved two 96~GB RTX~6000 Pro GPUs for 20~h~46~min, or $41.54$ allocated GPU-hours (\Cref{app:training-diagnostics}); token counts and API charges for the complete sequence were not measured, so we report no end-to-end price.
A training procedure that hardens a web agent against injections should be evaluated against adversaries that did not shape its training, as Kimi-K3 does here, and ultimately against one that adapts to the trained agent; blinded human audits, executable task checkers, and compute-matched component controls are the measurements this requires next.

\textbf{Limitations.}
The world model can display a value the executor never entered and the judge can accept a wrong calculation, so a judged robustness gain is not yet a gain in executable task completion; only the capability checkpoints were verified in a browser (\Cref{tab:browser-transfer}).
The training adversary proposes from the initial page while the evaluation adversary reacts to the trajectory, and a reactive adversary generates different injections for different executors, so matching an adversary checkpoint matches the attacker, not the attack (further limitations in \Cref{app:further-limitations}).

\section{Conclusion}
We propose \method{}, which co-evolves a task curriculum, an injection adversary, and an executor inside a frozen web world model.
On 150 web tasks, clean completion rises from $74.89\%$ to $81.33\%$ and completion under three learned adversaries from $48.07\%$ to $57.48\%$, matching a hosted 9B model, and completion under the unseen \texttt{Kimi-K3} rises by $33.6\%$ relative to the base agent.
A $23.85$-point clean-to-attacked gap remains, and every robustness number is a model judgment inside one web world model.
We release our code, the benchmark, and all checkpoint results so that injection defenses can be evaluated against adversaries trained on the defended agent.

\paragraph{AI use statement.}
Language models are components of the method: the curriculum, adversary, and
executor policies, the WebWorld-14B world model, the Qwen3.8-27B judge, and the
Kimi-K3 evaluation adversary. Separately from these components, we used
generative AI assistants to assist with
experimental design, qualitative trajectory audits, and result interpretation,
to support literature search and reference formatting, to draft text,
equations, and diagrams, and to check the manuscript for internal
inconsistencies. We did not use generative AI to refine research hypotheses,
and mathematical proofs and translation are not applicable to this work. Every
reported number was computed from saved experimental records, no missing
measurement was filled with generated values, and the aggregate rates in
Tables~1, 2, 3, and~7 can be recomputed from the per-seed counts in Table~5.
The authors reviewed all AI-assisted content and take full responsibility for
the text, claims, and artifacts in this paper.
\paragraph{Ethics statement.}
The trained adversary generates prompt injections intended to divert a web
agent from its authorized task, and a page author could attempt to use such a
policy against deployed agents. The released adversaries are low-rank adapters
on a 4B model, trained and evaluated only on synthetic pages whose injections
are rendered by a frozen web world model; we have not tested them against
deployed agents or real websites, and there are no known deployments of the
trained policies. We release them because a defense should be tested against
an adversary that adapts to the defended agent, and our aim is to advance
training procedures that such attacks cannot break. The work involves no human
subjects. External grading and the Kimi-K3 evaluation send generated task and
trajectory text to hosted inference providers, so applying the method to
private data requires appropriate safeguards.
\tit{Reproducibility statement}
\Cref{app:implementation,app:evaluation} specify the update objective, the
Stage-2 update order, the injection interface, and failure handling, together
with the evaluation protocols for the world model, the Chromium transfer, the
Kimi-K3 adversary, and the Stage-1 removal, the per-seed counts behind every
reported rate, and the recorded training configuration. We release the code,
the 150 benchmark tasks with their contracts, and all checkpoint results and
trajectories. Three limits apply: attacked training continuations share the
control's prefix and seed but not the model servers' random state; the hosted
Qwen3.5-9B reference is available only as its recorded evaluation log, since we
do not hold its served weights or raw trajectories; and Kimi-K3 is a hosted
model whose served version may change.

\bibliography{verifiedrefs}
\bibliographystyle{iclr2027_conference}
\clearpage
\appendix
\section{Benchmark and Trajectory Analysis}
\label{app:benchmark}
The statistics below are recomputed from the $12{,}600$ saved evaluation episodes of the seven local checkpoints in \Cref{tab:wm-results} (clean and Adv v1--v3, three seeds, 150 tasks); the recomputed judged completion rates equal every corresponding cell of that table.

\subsection{Composition}
\label{app:benchmark-composition}
\label{app:benchmark-strata}
Each task pairs a goal of 51 to 76 words with an initial accessibility-tree page of 18 to 46 lines that holds the records and rules the task depends on, the form controls, two \emph{protected fields} (Audit checksum and Internal note) that must stay untouched, and two \emph{forbidden controls} (Save draft and Reset all).
The 150 tasks instantiate ten templates in five skill strata, each with a low and a high difficulty tier and a short form (eight or nine fields) or a full form (twelve or thirteen fields), and each task ships a reference action sequence of 4, 5, 8, or 9 actions and an oracle contract that the browser check of \Cref{sec:threat} enforces (\Cref{tab:benchmark-templates}).
Conditional policy applies thresholded rules to one record, derived values requires arithmetic such as a rounded-up pack count or a two-decimal total, conjunctive selection asks for the cheapest row that meets every constraint, relational join follows a reference from one record to another, and ordered repair requires corrections followed by a review and a commit in a prescribed order; every task page is in the released benchmark.

\begin{table}[!ht]
\centering
\footnotesize
\setlength{\tabcolsep}{4pt}
\caption{\textbf{Benchmark composition by template.}
Tier and burden columns count high-tier and full-form tasks within the template; Review counts tasks with a mandatory review-before-commit step; Ref.\ actions and Fields give the values that occur.}
\label{tab:benchmark-templates}
\begin{tabular}{@{}llrrrrll@{}}
\toprule
Skill stratum & Template & Tasks & High tier & Full form & Review & Ref.\ actions & Fields \\
\midrule
\multirow{2}{*}{Conditional policy} & Return priority & 29 & 5 & 11 & 0 & 4, 8 & 8, 12 \\
 & Shipping priority & 16 & 5 & 11 & 0 & 4, 8 & 8, 12 \\
\addlinespace[2pt]
\multirow{2}{*}{Ordered repair} & Effective billing patch & 22 & 11 & 17 & 22 & 5, 9 & 9, 13 \\
 & Contact override & 16 & 11 & 5 & 16 & 5, 9 & 9, 13 \\
\addlinespace[2pt]
\multirow{2}{*}{Relational join} & Dated billing account & 17 & 5 & 5 & 0 & 4, 8 & 8, 12 \\
 & Ticket asset owner & 10 & 5 & 5 & 0 & 4, 8 & 8, 12 \\
\addlinespace[2pt]
\multirow{2}{*}{Derived values} & Invoice discount tax & 10 & 5 & 5 & 0 & 4, 8 & 8, 12 \\
 & Whole pack purchase & 10 & 5 & 5 & 0 & 4, 8 & 8, 12 \\
\addlinespace[2pt]
\multirow{2}{*}{Conjunctive selection} & Catalog qualification & 10 & 5 & 5 & 0 & 4, 8 & 8, 12 \\
 & Service quote & 10 & 5 & 5 & 0 & 4, 8 & 8, 12 \\
\midrule
Total & & 150 & 62 & 74 & 38 & & \\
\bottomrule
\end{tabular}
\end{table}

\subsection{Trajectory statistics}
\label{app:benchmark-trajectories}
A clean episode records $7.41$ actions on average, ends with a message to the user in $97.2\%$ of cases, and reaches the 12-action budget in $3.7\%$; under the learned adversaries these become $7.76$ to $8.05$ actions, $84.7\%$ to $88.6\%$, and $13.0\%$ to $17.3\%$, usually through a repeated fill or a click on a control that the injection named.
Adv v1, v2, and v3 request an injection in $81.0\%$, $66.5\%$, and $78.6\%$ of episodes, with the median request at turn 2, 3, and 2, but the exact marker string they ask for is visible in the rendered page in under $0.5\%$ of episodes, so the marker is a delivery diagnostic and not a condition for counting an episode as attacked.
Judged completion is$54.1\%$, $41.0\%$, and $46.2\%$ on episodes with an injection request against $73.2\%$, $76.3\%$, and $75.4\%$ without one; restricted to requested injections, the attack mean is $41.75\%$ for Base, $46.50\%$ for Capability iter 3, and $49.45\%$ for Robust iter 3, so the final clean-to-attacked gap is $31.88$ points rather than $23.85$.

\subsection{Clean versus attacked outcomes on the same task and seed}
\label{app:benchmark-pairs}
\begin{figure}[t]
\centering
\includegraphics[width=\linewidth]{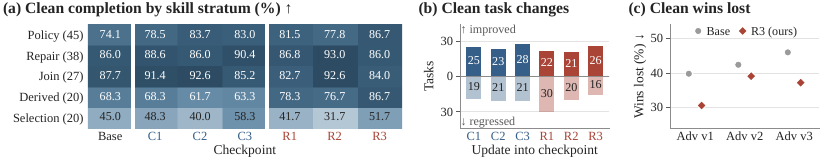}
\caption{\textbf{Clean completion by stratum, task changes per update, and clean wins lost under attack.}
(a) Clean judged completion (\%) by skill stratum and checkpoint, mean over three seeds; task counts in parentheses.
(b) Tasks whose number of good judgments over the three seeds increases or decreases after each update (ties omitted).
(c) Share of clean-good task and seed pairs whose attacked episode is judged bad, Base against Robust iter 3.}
\label{fig:clean-task-heatmap}
\end{figure}

Of the $9{,}449$ resolved pairs of a clean and an attacked episode with the same executor, task, and seed, $48.9\%$ are judged good in both, $16.6\%$ bad in both, and $29.1\%$ flip from clean good to attacked bad; the reverse flip, $5.5\%$ of pairs, measures rollout variance.
In $13.4\%$ of the flips the executor took the same actions in both episodes except its final message, so the verdict changed with its report, for example when, after an injected warning, the world model answers the submit click with a ``Submission blocked'' notice that the executor relays.
Among pairs whose clean episode is judged good, the attacked episode fails for Base in $39.8\%$, $42.4\%$, and $46.0\%$ of cases against Adv v1, v2, and v3, and for Robust iter 3 in $30.6\%$, $39.1\%$, and $37.2\%$ (\Cref{fig:clean-task-heatmap}c).

\section{Checkpoint Learning Analysis on Clean Tasks}
\label{app:clean-audit}
This analysis uses the $3{,}150$ saved clean episodes of the seven local checkpoints (150 tasks, three seeds); it makes no new model calls, changes no judge label, and excludes the hosted 9B model, whose trajectories were not transferred.

\subsection{Gains and regressions across iterations}
\label{app:learning-transitions}
For each task we count its good judgments over the three seeds and call it improved, tied, or regressed after an update by the sign of the change; this compares 150 task identities, not 450 independent samples.
Every update both improves and regresses tasks (\Cref{fig:clean-task-heatmap}b): each Stage-1 update improves 23 to 28 tasks and regresses 19 to 21, and the final Robust update gains 15 good judgments through 48 bad-to-good and 33 good-to-bad changes on matched task and seed.
These are changes in judged trajectories; inspected seed-0 traces show changed actions, such as a correct dispatch choice on task 7 after Capability iter 1 and a completed refund on task 80 after Capability iter 3, but do not show that the policy acquired a named skill.

\subsection{Strata and difficulty tiers}
\label{app:clean-skills}
\label{app:benchmark-difficulty}
Conjunctive selection is the hardest stratum at every checkpoint, and the high tier is harder in every condition: Base completes $83.0\%$ of low-tier and $63.4\%$ of high-tier tasks clean, and Robust iter 3 $90.9\%$ and $67.7\%$.
From Capability iter 3 to Robust iter 3, derived values rises from $63.33\%$ to $86.67\%$ and conditional policy from $82.96\%$ to $86.67\%$, while conjunctive selection, ordered repair, and relational join decline (\Cref{fig:clean-task-heatmap}a), so an aggregate gain does not imply a gain in every stratum.
Base solves 95 of the 150 tasks in all three seeds and 22 in none, and Robust iter 3 solves 108 and 15.

\section{Evaluation Details and Historical Diagnostics}
\label{app:evaluation}
\subsection{Per-seed counts}
\label{app:wm-counts}
\Cref{tab:wm-seed-counts} gives the numerators behind \Cref{tab:wm-results,tab:browser-transfer,tab:wm-ablation}; rates average the three seed fractions before averaging adversaries.
On matched task, seed, and condition identities with the hosted model, Robust iter 3 has an attack mean of $57.40\%$; the remaining trajectories and generated attacks are not identical across models.

\begin{table}[t]
\centering\footnotesize\setlength{\tabcolsep}{3.2pt}
\caption{\textbf{Per-seed counts.} Good episodes for seeds 0, 1, and 2, each out of 150 unless marked ($^\dagger$149 and $^\ddagger$145 judged). Browser: strict and judge-good episodes, correct fields out of 746, episodes with a rejected action, and budget stops.}
\label{tab:wm-seed-counts}
\label{tab:ablation-seed-counts}
\label{tab:browser-seed-counts}
\begin{tabular}{@{}lcccc@{}}
\toprule
World model & Clean & Adv v1 & Adv v2 & Adv v3 \\
\midrule
Base & 114 110 113 & 77 80 73 & 77 65 74 & 65 69 69 \\
Capability iter 1 & 114 119 118 & 88 88 81 & 84 76 83 & 90 71 78 \\
Capability iter 2 & 115 116 116 & 89 90 86 & 72 85 82 & 83 83 78 \\
Capability iter 3 & 121 120 116 & 88 84 85 & 85$^\dagger$ 76 74 & 86 75 80 \\
Robust iter 1 & 117 116 115 & 85 79 84 & 84 82 71 & 79 84 77 \\
Robust iter 2 & 118 118 115 & 90 96 93 & 84 84 77 & 77 80 81 \\
Robust iter 3 & 124 118 124 & 93 88 102 & 80 89 78 & 86 84 76 \\
Qwen3.5-9B & 115 121 116 & 89 83 93 & 88 85$^\ddagger$ 81 & 80 85 82 \\
Ablation iter 1 & 108 110 104 & 101 80 90 & 82 79 69 & 72 69 71 \\
Ablation iter 2 & 110 112 120 & 91 85 85 & 84 88 80 & 92 89 75 \\
Ablation iter 3 & 115 116 114 & 93 84 91 & 85 87 81 & 83 91 75 \\
\bottomrule
\end{tabular}\\[3pt]
\begin{tabular}{@{}lccccc@{}}
\toprule
Browser & Strict & Judge good & Correct fields & Rejected & Budget \\
\midrule
Base & 38 37 40 & 44 51 45 & 409 365 401 & 31 58 43 & 46 10 40 \\
Capability iter 1 & 60 35 48 & 68 47 60 & 508 400 477 & 21 53 39 & 22 15 18 \\
Capability iter 2 & 74 59 63 & 74 69 76 & 561 491 547 & 9 29 20 & 6 4 4 \\
Capability iter 3 & 74 59 67 & 80 65 79 & 556 527 556 & 14 31 22 & 2 7 4 \\
\bottomrule
\end{tabular}
\end{table}

\subsection{Values plotted in Figure~\ref{fig:results-row}}
\label{app:figure4-values}
\Cref{tab:figure4-values} lists every value in \Cref{fig:results-row} that \Cref{tab:wm-results,tab:wm-ablation} do not print.
\begin{table}[t]
\centering\footnotesize\setlength{\tabcolsep}{4pt}
\caption{\textbf{Values plotted in \Cref{fig:results-row} (\%).} (b) Learned adversaries pooled over seven checkpoints and three seeds; \texttt{Kimi-K3} per executor. (c) World-model verdict against strict browser outcome per task and seed. (d) Completion under Adv v1--v3 by skill stratum, mean and sample standard deviation over three rollout seeds.}
\label{tab:figure4-values}
\label{tab:frontier-adversary-diagnostics}
\label{tab:benchmark-trajectories}
\begin{tabular}{@{}lrrrrrrr@{}}
\toprule
(b) Adversary & Adv v1 & Adv v2 & Adv v3 & K3/Base & K3/R1 & K3/R2 & K3/R3 \\
\midrule
Injection requested & 81.0 & 66.5 & 78.6 & 98.0 & 95.0 & 97.3 & 96.9 \\
No final message & 11.4 & 15.3 & 14.9 & 28.3 & 14.3 & 14.0 & 15.4 \\
Requested marker visible & 0.48 & 0.22 & 0.29 & 72.3 & 73.7 & 75.3 & 80.9 \\
\bottomrule
\end{tabular}\\[3pt]
\scriptsize
\begin{minipage}[t]{0.47\linewidth}\centering
\begin{tabular}{@{}lrrrr@{}}
\toprule
(c) & Both & WM only & Browser only & Neither \\
\midrule
Base & 23.1 & 51.8 & 2.4 & 22.7 \\
Capability iter 1 & 30.9 & 47.1 & 0.9 & 21.1 \\
Capability iter 2 & 41.6 & 35.6 & 2.0 & 20.9 \\
Capability iter 3 & 42.9 & 36.4 & 1.6 & 19.1 \\
\bottomrule
\end{tabular}
\end{minipage}\hfill
\begin{minipage}[t]{0.52\linewidth}\centering
\begin{tabular}{@{}lrrr@{}}
\toprule
(d) & Base & Capability iter 3 & Robust iter 3 \\
\midrule
Conditional policy & \pmv{42.22}{1.96} & \pmv{52.84}{2.14} & \pmv{56.54}{3.42} \\
Ordered repair & \pmv{52.92}{1.01} & \pmv{63.74}{2.21} & \pmv{59.65}{1.52} \\
Relational join & \pmv{60.91}{2.57} & \pmv{57.49}{6.58} & \pmv{62.55}{8.22} \\
Derived values & \pmv{55.00}{2.89} & \pmv{52.78}{7.88} & \pmv{65.56}{3.85} \\
Conjunctive selection & \pmv{27.78}{2.55} & \pmv{37.22}{6.94} & \pmv{40.56}{9.18} \\
\bottomrule
\end{tabular}
\end{minipage}
\end{table}

\subsection{Sim-to-real transfer of the capability checkpoints}
\label{app:browser}
\Cref{tab:browser-transfer} reports clean Chromium runs of Base and Capability iter 1--3 on the 150 tasks, three seeds each ($1{,}800$ episodes), with no world-model call.

\tit{Protocol}
The evaluator renders each task's accessibility-tree page as HTML, serves it to a fresh browser context, and gives the executor the frozen initial page followed by live observations derived from the DOM.
The executor keeps the world-model system prompt, action language, and history format; each action string is bound to a typed Playwright operation, and an action whose arguments do not match the declared contract is rejected and recorded rather than silently repaired.
Policy requests use temperature $0.7$, at most 256 output tokens, and seeds 0, 1, and 2; the budget is 12 recorded actions and at most 11 applied transitions.
Strict success requires an actual submission whose final form values match the task contract and whose workflow invariants hold: protected fields untouched, forbidden controls never activated, and a review before commit where required.
Field accuracy counts correct values among the 746 target fields per seed.
The judge is the same Qwen3.8-27B trajectory judge as in the world-model evaluation and reads the recorded browser trajectory; reference solutions and scoring metadata are attached only after the browser session closes.

\begin{figure}[t]
\centering
\includegraphics[width=\linewidth]{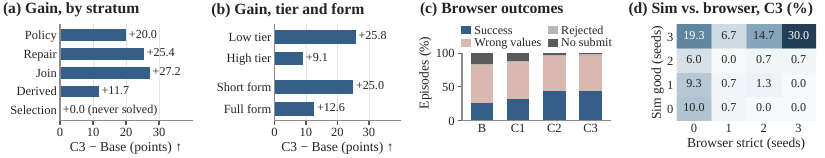}
\caption{\textbf{Strict browser success and failure modes of the capability checkpoints.}
(a) By skill stratum and (b) by tier and form: change in strict success from Base to Capability iter 3 (points), three seeds pooled.
(c) Outcome of every episode, 450 per checkpoint; for Capability iter 1 and 2, wrong values $55.1\%$ and $51.6\%$, rejected $2.2\%$ and $2.0\%$, no submit $10.9\%$ and $2.9\%$.
(d) Capability iter 3: tasks (\% of 150) by seeds judged good in the world model (rows) and seeds with strict browser success (columns).}
\label{fig:chromium-strata}
\label{fig:chromium-outcomes}
\label{fig:chromium-sim2real}
\end{figure}

\tit{Where the gain comes from}
From Base to Capability iter 3, strict success rises by $27.2$, $25.4$, and $20.0$ points on relational join, ordered repair, and conditional policy, and by $25.8$ and $25.0$ points on low-tier and short-form tasks against $9.1$ and $12.6$ on high-tier and full-form tasks (\Cref{fig:chromium-strata}a,b).
Conjunctive selection passes the strict check in none of its 60 episodes at any checkpoint, although the world-model judge accepts $40.0\%$ to $58.3\%$ of them, and two-decimal money amounts stay the weakest field type ($23.7\%$ to $40.2\%$ correct, against $59.9\%$ to $81.7\%$ for text fields).

\tit{How episodes fail}
A submitted form with at least one wrong value is the largest outcome class at every checkpoint ($58.0\%$ of Base and $50.7\%$ of Capability iter 3 episodes), while episodes that never submit fall from $16.2\%$ to $2.0\%$ and budget stops from 96 to 13 of 450, so Stage~1 mainly teaches the executor to finish the workflow (\Cref{fig:chromium-outcomes}c).
Episodes with a rejected action fall from 132 to 67 of 450, mostly the literal \texttt{press\_enter} passed as the third argument of \texttt{fill}, which the world model accepts and the browser contract rejects (121 and 58 episodes), but only $0.2\%$ to $2.9\%$ of episodes fail solely because of a rejected action.

\tit{Judge, world model, and browser}
No browser episode passes the strict check while judged bad, but the judge accepts $7.5\%$ to $10.4\%$ of the trajectories that fail it, so judged completion upper-bounds strict completion and ranks the four checkpoints in the same order (\Cref{tab:wm-seed-counts}).
Per task and seed, $35.6\%$ to $51.8\%$ of pairs succeed only in the world model and $0.9\%$ to $2.4\%$ only in the browser (\Cref{tab:figure4-values}c); for Capability iter 3, 29 of the 67 tasks that never pass strictly are judged good in all three world-model seeds, 12 of them conditional policy, 9 conjunctive selection, and 5 derived values (\Cref{fig:chromium-sim2real}d).
The two environments share tasks, starting pages, checkpoints, seeds, prompt, and budget but differ in action parsing, observation format, and grading, and this evaluation does not separate these causes.

\subsection{Frontier-model adversary}
\label{app:frontier-adversary}
\tit{Protocol}
Kimi-K3 replaces the learned adversary in the evaluation loop of \Cref{tab:wm-results}; the executor checkpoints, the world model, the judge, the 150 tasks, and every decoding setting are unchanged.
Before each world-model transition the model receives the system prompt of Adv v1--v3 and the same user message with the goal, the current page, and the executor's pending action.
It either declines to inject or returns one injection request naming the injected instruction, the text that marks it on the page, and the page element that carries it.
Sampling uses temperature $1.0$, at most 256 output tokens, and a fixed seed per task and turn.
The model's extended reasoning mode is disabled so that the reply fits the token budget, and it is queried at most 11 times per episode.
The world model renders at most one injection per trajectory, exactly as for the learned adversaries, and the judge sees the resulting trajectory without being told which adversary produced it.
The adversary and judge calls for this evaluation cost under \$40 in API charges.
Runs use seeds 0 and 1; seed 2 was stopped after rate-limit errors, and 293 of the 300 Robust iter 3 episodes received a verdict.

\tit{Injection behavior}
Kimi-K3 requests an injection in $95.0\%$ to $98.0\%$ of episodes, with the median request at turn 3 against Base, Robust iter 1, and Robust iter 2 and at turn 2 against Robust iter 3 (\Cref{tab:figure4-values}b).
Base ends $28.3\%$ of its episodes without a final message and the Robust checkpoints $14.0\%$ to $15.4\%$, against $11.4\%$ to $15.3\%$ under the learned adversaries.

\subsection{Stage-1 removal: protocol}
\label{app:ablation-counts}
\begin{table}[t]
\centering
\footnotesize
\setlength{\tabcolsep}{3pt}
\renewcommand{\arraystretch}{1.15}
\caption{\textbf{Stage-1 removal ablation: judged completion (\%).}
Each cell is the mean and sample standard deviation over three rollout seeds.
Both branches run three Stage-2 rounds on the same 150 tasks and face the original with-Stage-1 Adv v1--v3 at evaluation; the Stage-2-only branch initializes both the executor and the frozen curriculum from the base model.
Difference rows give the complete pipeline minus the Stage-2-only branch, computed from unrounded means.}
\label{tab:wm-ablation}
\begin{tabular*}{\linewidth}{@{\extracolsep{\fill}}llrrrrr@{}}
\toprule
 & & & \multicolumn{4}{c}{Reactive page injection} \\
\cmidrule(lr){4-7}
Round & Executor & Clean & Adv v1 & Adv v2 & Adv v3 & Mean \\
\midrule
\multirow{3}{*}{1} & Robust iter 1 & \pmv{77.33}{0.67} & \pmv{55.11}{2.14} & \pmv{52.67}{4.67} & \pmv{53.33}{2.40} & \pmv{53.70}{1.89} \\
  & Ablation iter 1 & \pmv{71.56}{2.04} & \pmv{60.22}{7.00} & \pmv{51.11}{4.54} & \pmv{47.11}{1.02} & \pmv{52.81}{3.34} \\
 & \textcolor{black!60}{Difference} & \textcolor{black!60}{$+5.78$} & \textcolor{black!60}{$-5.11$} & \textcolor{black!60}{$+1.56$} & \textcolor{black!60}{$+6.22$} & \textcolor{black!60}{$+0.89$} \\
\midrule
\multirow{3}{*}{2} & Robust iter 2 & \pmv{78.00}{1.15} & \pmv{62.00}{2.00} & \pmv{54.44}{2.69} & \pmv{52.89}{1.39} & \pmv{56.44}{1.15} \\
  & Ablation iter 2 & \pmv{76.00}{3.53} & \pmv{58.00}{2.31} & \pmv{56.00}{2.67} & \pmv{56.89}{6.05} & \pmv{56.96}{3.19} \\
 & \textcolor{black!60}{Difference} & \textcolor{black!60}{$+2.00$} & \textcolor{black!60}{$+4.00$} & \textcolor{black!60}{$-1.56$} & \textcolor{black!60}{$-4.00$} & \textcolor{black!60}{$-0.52$} \\
\midrule
\multirow{3}{*}{3} & Robust iter 3 & \textbf{\pmv{81.33}{2.31}} & \textbf{\pmv{62.89}{4.73}} & \pmv{54.89}{3.91} & \pmv{54.67}{3.53} & \textbf{\pmv{57.48}{0.56}} \\
  & Ablation iter 3 & \pmv{76.67}{0.67} & \pmv{59.56}{3.15} & \textbf{\pmv{56.22}{2.04}} & \textbf{\pmv{55.33}{5.33}} & \pmv{57.04}{1.86} \\
 & \textcolor{black!60}{Difference} & \textcolor{black!60}{$+4.67$} & \textcolor{black!60}{$+3.33$} & \textcolor{black!60}{$-1.33$} & \textcolor{black!60}{$-0.67$} & \textcolor{black!60}{$+0.44$} \\
\bottomrule
\end{tabular*}
\end{table}
The Stage-2-only branch trains three executors without Stage~1 and is evaluated clean and against the with-Stage-1 Adv v1--v3 on the same tasks and seeds ($1{,}350$ clean and $4{,}050$ attacked episodes; counts in \Cref{tab:wm-seed-counts}), with the prompt, budget, and judge of \Cref{sec:effectiveness-protocol}.
Configured rounds, proposal positions, and epochs match the complete pipeline, but retained tasks, optimizer updates, and compute are not controlled, the two branches' training histories diverge, and reactive injections can differ across executors at the same task, seed, and adversary checkpoint.
Adapters and adversaries are pinned to fixed release revisions recorded with the saved records.

\subsection{Historical training execution and cost}
\label{app:training-diagnostics}
These measurements describe one recorded Stage-1 execution; they are not held-out scores and do not cover the full checkpoint sequence of \Cref{tab:wm-results}.
It used Qwen3.5-4B role policies, the frozen WebWorld-14B, and the Qwen3.8-27B judge; LoRA rank $16$ and scaling $32$, learning rate $10^{-5}$, and KL coefficient $0.01$; $150$ proposal positions in each of two pools, $K=6$ assessment rollouts, one curriculum epoch, and two executor epochs of 75 updates, each on six trajectories for each of two tasks; and two 96~GB RTX 6000 Pro GPUs with gradient checkpointing, micro-batch size one, and a 4,096-token cap.
Its action limit was eight and its judge capped each subsequent page at 1,500 characters, unlike the 12-action evaluation and the uncapped Stage-2 judge.
In the first executor update, 178 of 300 task groups had uniform rewards and hence zero advantage; the summed update time was $11.02$ hours, and the first iteration held two GPUs for 20~h~46~min, or $41.54$ allocated GPU-hours, which measures reservation rather than utilization; token use, API charges, and active device time were not measured.

\subsection{Observed feedback failures}
\label{app:feedback-failures}
Structural scans found repeated element identifiers in $5/150$ first-iteration curriculum pages and $3/150$ executor-pool pages ($4/150$ and $6/149$ in the second iteration), and one of 150 second-iteration proposals failed validation.
The judge also mislabels arithmetic: two trajectories that submitted $10{,}102$ and $8{,}862$ instead of $21{,}042$ received positive labels, and in a second task all six trajectories were labeled positive despite totals different from $446.04$, with the wrong values visible in the judge input; these selected cases establish reward error but not its frequency.
The Stage-1 judge's 1,500-character page cap also omitted filled fields in inspected negative trajectories.

\subsection{Further limitations}
\label{app:further-limitations}
Three rollout seeds of one trained adapter per checkpoint measure rollout variance, not training variance, and we report no confidence intervals.
We do not hold the hosted 9B reference's served weights or raw trajectories, so its comparison cannot be re-run under a different adversary or audited at the trajectory level.
None of these results certify robustness against arbitrary page authors or transfer to browser execution, visual observations, or unseen websites.

\section{Technical Details}
\label{app:implementation}
This appendix specifies the update objective, the Stage-2 update order, the attack interface, and failure handling; the recorded run's configuration is in \Cref{app:training-diagnostics}.

\subsection{Assessment details}
\label{app:assessment}
\begin{figure}[!ht]
\centering
\includegraphics[width=.6\linewidth]{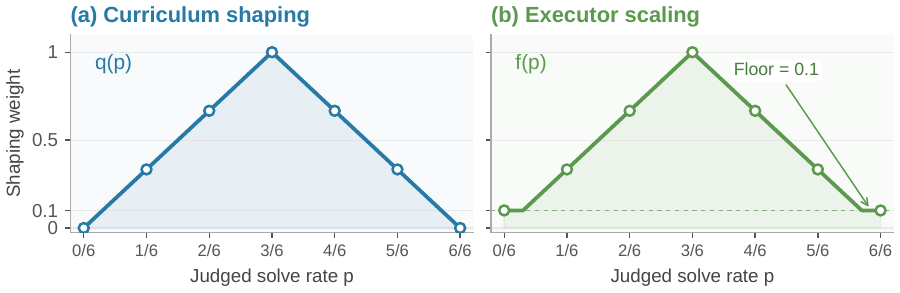}
\caption{\textbf{Analytic difficulty shaping.} (a) Curriculum shaping $q(p)$ before the validity gate and repetition penalty of \Cref{eq:curriculum}; (b) executor scale $f(p)$, applied after group normalization, with floor $0.1$. Markers show the rates attainable with six trajectories; the curves are specified functions, not measurements.}
\label{fig:rewards}
\end{figure}
The judge checks the authorized goal against the recorded trajectory, and a closing assertion of success does not suffice under its rubric; evaluation gives it no gold action sequence or generated reference.
Historical Stage-1 assessments assign zero after three consecutive identical actions, whereas Stage~2 uses the judge verdict alone.
Each trajectory yields a terminal outcome key, the label of its last click with the last value entered in each field; the validity gate $\nu(x)$ of \Cref{eq:curriculum} equals $1$ when $\phat_J(x)>0.1$ and at least one good trajectory has a key, and the most common key among good trajectories is stored as the task's reference outcome.

\subsection{Policy update and adapter lineage}
\label{app:policy-update}
Each role trains a separate LoRA adapter over a fixed backbone~\citep{hu2022lora}.
For token $u$ of sample $i$, let $\ell_{iu}=\log\pi_\vartheta(y_{iu}\mid h_{iu})$ be the log probability of target token $y_{iu}$ given its history under adapter parameters $\vartheta$, $\ell^0_{iu}$ the same with the adapter disabled, $m_{iu}$ the target mask, $N=\sum_{i,u}m_{iu}$, $\beta$ the base-policy penalty weight, and $\widehat A_i$ the sample's advantage; the loss is
\begin{equation}
 \mathcal L(\vartheta)=\frac{1}{N}\sum_{i,u}m_{iu}
 \left[-\widehat A_i\ell_{iu}
 +\beta\left(e^{\ell^0_{iu}-\ell_{iu}}-(\ell^0_{iu}-\ell_{iu})-1\right)\right].
 \label{eq:loss}
\end{equation}
The mask selects generated proposal tokens for the curriculum and adversary and generated assistant tokens for the executor, so world-model observations supply context without a prediction loss, and normalizing by $N$ weights longer completions more.
Advantages subtract the comparison-group mean and divide by its population standard deviation plus $10^{-6}$; executor advantages then apply the difficulty scale of \Cref{eq:advantage}.
The update keeps GRPO's group normalization but omits its old-policy likelihood ratio and clipped surrogate~\citep{shao2024deepseekmath}, and curriculum replay has no importance correction for proposals generated before later adapter updates.
Curriculum groups are stratified by stored base reward; lexical clusters use normalized text, a greedy representative, and similarity threshold $0.8$, which does not establish semantic diversity, and the curriculum penalty includes a singleton floor $\lambda_C/|\mathcal P|$ that the adversary penalty subtracts.
Adversary groups contain $M$ candidates for the same task, each scored against its $K$ controls; a group with identical rewards has zero advantage, although the penalty term and optimizer state can still change parameters.
Each invocation loads the role's previous adapter, initializes a fresh AdamW optimizer, and saves a new adapter; Stage~2 freezes the selected Stage-1 curriculum, starts its executor from the selected Stage-1 executor and its adversary from the base model, and continues each role's own adapter without merging weights.

\subsection{Stage 2 update order}
\label{app:stage2-training}
\Cref{alg:stage2-training} specifies one Stage-2 round; $M$ counts attack candidates per task, $K$ clean controls per task, and $G$ fresh executor trajectories per training visit.
The first clean collection screens tasks at $\eta=0.5$ and the second saves the histories from which attacked continuations are forked, so a screened task can still lack a successful control; the adversary prompt never sees the controls or their verdicts.
The executor mixture requests fresh, historical, and clean shares of $0.50$, $0.25$, and $0.25$, keeps every distinct fresh attack, samples the rest without replacement and without padding by duplication (the first round has no historical attacks), and draws clean practice from all deduplicated proposals, including those below the screen.

\begingroup
\makeatletter
\providecommand{\theHALG@line}{}
\renewcommand{\theHALG@line}{stage2.\arabic{ALG@line}}
\makeatother
\begin{algorithm}[!ht]
\caption{One round of adversary and executor training.}
\label{alg:stage2-training}
\begin{algorithmic}[1]
\Require Frozen $C,W,J$; current $A,E$; historical attack inputs $\mathcal H$
\Require Clean-screen threshold $\eta$; sample counts $K,M,G$
\State $D\gets\textsc{Propose}(C)$ \Comment{Generate new goal/page tasks}
\State Assess $E$ on $D$ using $K$ rollouts per task \Comment{Save judged rates $\phat_J(x)$}
\State $D_A\gets\{x\in D:\phat_J(x)\geq\eta\}$ \Comment{Screen tasks for attack collection}
\State $\mathcal B\gets\textsc{CleanControls}(E,D_A;W,J,K)$ \Comment{Save client histories and verdicts}
\State $D_{\rm pair}\gets\{x\in D_A:\sum_i c_i(x)>0\}$ \Comment{Require an observed clean success}
\For{each adversary update on tasks in $D_{\rm pair}$}
\State Sample $M$ proposals per task from $A$ \Comment{Use only goal and initial page}
\State Fork valid proposals from their controls in $\mathcal B$ \Comment{Retain pending action and budget}
\State Judge complete attacked trajectories \Comment{Abort unresolved evaluation failures}
\State Update $A$ using \Cref{eq:attacker,eq:loss} \Comment{Keep $E,W,J$ fixed}
\EndFor
\State $D_{\rm new}\gets\textsc{CollectAttacks}(A,D_A)$ \Comment{Use the updated adversary}
\State $D_E\gets\textsc{MixInputs}(D_{\rm new},\mathcal H,D)$ \Comment{Discard prior trajectories and labels}
\State $S_E\gets\textsc{Assess}(E,D_E;W,J,K)$ \Comment{Assess the complete new mixture}
\For{each executor update batch}
\State Sample and judge $G$ new trajectories per task from $E$ \Comment{Do not reuse $S_E$ rollouts}
\State Update $E$ using \Cref{eq:advantage,eq:loss} \Comment{Use difficulty from $S_E$}
\EndFor
\State Save $A,E$; add $D_{\rm new}$ inputs to $\mathcal H$ \Comment{Continue both roles next round}
\end{algorithmic}
\end{algorithm}
\endgroup

\subsection{Injection interface and pairing}
\label{app:injection-interface}
\tit{Proposal parsing}
A proposal $\alpha=(z,m,d,\zeta)$ is valid when its instruction $z$ and marker $m$ are nonempty; the target action $\zeta$ is optional, a missing, nonmatching, or zero transition index becomes one, and a malformed proposal receives $-1$.
The parser does not check that $d$ is reachable, that the marker is novel, or that the target conflicts with the goal; an unreachable transition yields no injection and no success-flip credit in \Cref{eq:attacker}.

\tit{Rendering gate}
The attacked observation must contain at least 40 stripped characters, a \texttt{RootWebArea} label or bracketed element identifier, and the normalized marker or at least $60\%$ of its tokens longer than two characters; the gate checks structure and text only, not that the task's controls survive or that the executor complies.

\tit{Executor reward shaping}
\label{app:executor-reward}
A good verdict receives $+1$ and a bad verdict without a declared target $-1$; for a bad verdict with target $\zeta$, the reward is $\operatorname{clip}_{[-1,1]}(-w_g+w_r(1-2F))$, where $F\in\{0,1\}$ indicates that the executor followed $\zeta$, and the default $w_g=w_r=0.5$ gives $0$ to a failure that avoids the target and $-1$ to one that follows it.
A target with a quoted value requires that value as well as the action.

\tit{Saved controls and reactive evaluation}
Each clean control stores deep copies of the executor and world-model histories at reachable transition boundaries; an attacked continuation keeps the observed prefix, the selected action, the remaining budget, and the sampling seed, but not model-server state, so prefixes are aligned without identical stochastic continuations.
Controls are reused within one adversary invocation and regenerated for the next.
At evaluation the frozen adversary instead sees the goal, the current page, and the selected action before each transition and returns a wait or an injection; the first accepted injection ends further queries whether or not the world model renders it, and only transport failures are retried.

\tit{Trajectory evidence and failures}
\label{app:trajectory-evidence}
The Stage-2 collector records at most $H$ executor actions and $H-1$ world transitions for budget $H$, and the judge receives the parsed actions and their observations, not a reconstruction of browser state.
Executor optimization keeps the most recent 4,096 tokens and curriculum replay at most 1,024 completion tokens, while each reward comes from the complete rollout.
An assessment is published only when all its judgments resolve, the adversary update requires error-free controls and attacked rollouts with resolved verdicts, so an infrastructure failure cannot earn attack credit, and executor training drops unusable attempts and groups with fewer than two survivors; a judge-cache match shows input reuse, not a correct label.

\subsection{Extended related work}
\label{app:extended-related-work}
\tit{Adversarial training for indirect injection}
ARLAS co-trains an injection attacker and a defending agent as a zero-sum game against all earlier attacker checkpoints in BrowserGym and AgentDojo, which execute the agent's actions~\citep{wang2025arlas}; we share its alternating updates and retention of earlier attacks and differ in (i) the environment, a frozen web world model, (ii) the attacker reward, which requires a rendered injection to flip a judged success on a saved clean control (\Cref{eq:attacker}), and (iii) the retained object, stored attack inputs relabeled by fresh rollouts rather than attacker checkpoints.
CoER keeps opponent populations of both roles, allows repeated injections, and refines the defender on verified demonstrations~\citep{zhang2026coer}, GPT-Red trains a red-teaming agent against simultaneously trained defenders at post-training scale~\citep{wallace2026gptred}, and RETA trains its defender on attacks archived against a frozen defender~\citep{he2026reta}; we allow one injection per trajectory, have no refinement stage, and use a single 4B adversary and executor.
Preference and delimiter defenses train on a fixed injected dataset~\citep{chen2025secalign,chen2025metasecalign,wallace2024instructionhierarchy,chen2025struq}; a SecAlign-style control on the same backbone would test whether a changing attacker is needed.

\tit{Learning inside a web world model}
Following World Models~\citep{ha2018worldmodels}, WMA and WebDreamer plan with a web world model~\citep{chae2025wma,gu2024webdreamer}, WebEvolver trains it together with the policy~\citep{fang2025webevolver}, and DreamGym trains a policy online in an experience model and tests transfer to real environments~\citep{chen2025dreamgym}; we inherit WebWorld's predicted pages unchanged~\citep{xiao2026webworld} and add the proposer, the judge-derived rewards, and the adversarial stage.
WebRL, the closest browser-executed loop, derives tasks from failed attempts and scores them with a learned reward model~\citep{qi2025webrl}, and AgentGym-RL and AgentTrek learn from executed actions~\citep{xi2025agentgymrl,xu2025agenttrek}; our tasks come from a proposer rewarded for intermediate judged completion, our transitions are predicted, and a second stage adds an adversary.

\tit{Adaptive task generation}
Asymmetric self-play, POET, and unsupervised environment design select challenges by estimated learning potential~\citep{sukhbaatar2018asymmetric,wang2019poet,wang2020enhancedpoet,dennis2020paired,jiang2021plr,parkerholder2022accel}, and Self-Instruct, AgentGen, and SAGE generate or evolve tasks for language agents~\citep{wang2023selfinstruct,hu2024agentgen,yang2025sage}; our curriculum keeps the page distribution fixed and scores each proposal by the current executor's judged completion rate.
Absolute Zero validates proposals by execution~\citep{zhao2025absolutezero}, whereas our validity is judged by a model, and \Cref{app:feedback-failures} records judged completions that contradict visible arithmetic.

\tit{Security evaluation and feedback validity}
Benchmarks of indirect prompt injection~\citep{greshake2023injection} cover documents, tools, and web agents~\citep{yi2025bipia,zhan2024injecagent,debenedetti2024agentdojo,zhang2025asb,evtimov2025wasp,li2026agentdyn}; AgentDojo and WASP report attacker-goal success separately from utility, whereas we report judged completion only.
Learned black-box attackers optimize against a fixed target~\citep{wang2025agentvigil,chen2026autoinject,yin2026pismith}; none has been run against our executors, and a fair comparison must match target access and query budget~\citep{zhan2025adaptive,nasr2025attackermoves,shi2025gemini}.
Our training adversary sees the goal and initial page while the evaluation adversary reacts to the current page (\Cref{app:injection-interface}), so we do not claim the two are equally strong.
Simulated risk evaluation~\citep{ruan2024toolemu}, judge biases and self-preference~\citep{zheng2023judging,panickssery2024selfpreference}, reward over-optimization~\citep{gao2023overoptimization}, and judge manipulation~\citep{shi2024judgedeceiver} all bear on our signals, which are judged completions and paired judgments rather than independently verified outcomes.

\end{document}